\documentclass[a4paper]{spie}  
\usepackage{graphicx}
\usepackage{amssymb, amsmath, subfigure}
\providecommand{\norm}[1]{\lvert#1\rvert}

\title{Exploring tradeoffs in pleiotropy and redundancy using evolutionary computing} 

\author{Matthew J. Berryman\supit{a}, Wei-Li Khoo\supit{a}, Hiep Nguyen\supit{a}, Erin O'Neill\supit{b}, Andrew Allison\supit{a}, and Derek Abbott\supit{a}
\skiplinehalf
\supit{a}Centre for Biomedical Engineering and\\
School of Electrical and Electronic Engineering,\\
The University of Adelaide, SA  5005, Australia.\\
\supit{b}Dept. of Human Physiology,\\
University of Newcastle,\\
Callaghan, NSW 2308, Australia.}

\authorinfo{Send correspondence to Derek Abbott\\
E-mail: dabbott@eleceng.adelaide.edu.au, Telephone: +61 8 8303 5748}

\begin{document} 
\maketitle
\begin{abstract}
Evolutionary computation algorithms are increasingly being used to solve optimization problems as they have many advantages over traditional optimization algorithms. In this paper we use evolutionary computation to study the trade-off between pleiotropy and redundancy in a client-server based network. Pleiotropy is a term used to describe components that perform multiple tasks, while redundancy refers to multiple components performing one same task. Pleiotropy reduces cost but lacks robustness, 
while redundancy increases network reliability but is more costly, as together, pleiotropy and redundancy build flexibility and robustness into systems. Therefore it is desirable to have a network that contains a balance between pleiotropy and redundancy. We explore how factors such as link failure probability, repair rates, and the size of the network influence the design choices that we explore using genetic algorithms.
\end{abstract}

\section{INTRODUCTION}
Evolutionary computation involves using solution space search procedures inspired by biological evolution~\cite{WinstonAI}. These search procedures use ideas from biological evolution such as mating, fitness, and natural selection. Individuals undergo natural selection, whereby organisms with the most favorable traits are more successful in having offspring. Genetic algorithms (GAs) rely on describing systems in terms of their traits (or phenotype) and then a fitness function (or how well they reproduce). Then we can evolve better solutions (with a higher fitness function) by allowing transfer of hereditary characteristics (genes) to the next generation for fit functions. The idea of applying such biological concepts to evolutionary computing was originates with John Holland in his seminal paper on the topic of adaptive systems~\cite{JohnHolland}.

Evolutionary computational techniques such as genetic algorithms have many advantages over traditional optimization algorithms. Current optimization algorithms require many assumptions to be made about the problem, for example with gradient-based searches, the requirement is that the function be smooth and differentiable. Evolutionary algorithms require no such assumptions, only requiring a way of measuring the ``fitness'' of a solution~\cite{IntroGA}. With each succeeding generation, the algorithm tries to better fulfill the specifications described by the fitness function. The other advantage is adaptability to a changing problem. For example with traditional optimization procedures, any change in the specification or problem constraints requires solving the problem from the start. This is not necessary with evolutionary algorithms where one can continue the algorithm with a different set of constraints or solution using the current ``population'' or set of solutions~\cite{Fogel}. GAs can offer advantages over related techniques such as hill climbing~\cite{mitchell94when,mitchell97}.

Although there are a large number of applications of genetic algorithms to designing neural networks~\cite{neuralnet1,neuralnet2,neuralnet3}, there are very few devoted to designing computer or telecommunication networks~\cite{netga1,netga2}, and none of these explicitly capture the issue of pleiotropy. An alternative evolutionary approach, cellular automata, has been applied to pleiotropy versus redundancy tradeoffs in an organizational system~\cite{Teck}. Pleiotropy is a term used to describe components that perform multiple tasks~\cite{pleiotropy1,pleiotropy2}, while redundancy refers to multiple components performing one same task. Such pleiotropy and redundancy of components can be clearly seen in a client-server based network comprising of server nodes and client nodes. The conventional setup of such networks can have servers serving multiple clients, which is an example of pleiotropy, while clients can be connected to many servers, which is an example of redundancy. 

A typical engineering problem is to determine the optimal design solution or set of solutions, while maximizing efficiency. The main aim of the project is to use evolutionary computation algorithms to search for an optimal client-server network, which minimizes cost and maximizes reliability and flexibility by exploring the pleiotropy-redundancy search space. 
\section{MOTIVATION}
Any system whose fundamental drive is to either function over a period of time (as an entity in its own right), or reproduce entities of great, but not exact similarity, has to work within two opposing constraints.  First, it must maintain its integrity, whether over the time of existence of that individual system, or from one generation to the next.  Secondly, the system must be able to adapt to (and potentially use) change both within itself and that imposed upon it by the environment.  The use of pleiotropy and redundancy within biological systems allows a system to work within those constraints.

Biological systems provide the best and most adaptive examples of pleiotropy and redundancy~\cite{cytokine1,cytokine2}.  Intercellular messenger molecules such as cytokines may act as links between nodes (cells)~\cite{CoussensWerb}.  If it were possible to understand how pleiotropy and redundancy worked within the cytokine networks, then manipulation of disease states would be possible~\cite{Mann,Palladino}.

\begin{figure}[hptb]
  \centering{\resizebox{6cm}{!}{\includegraphics{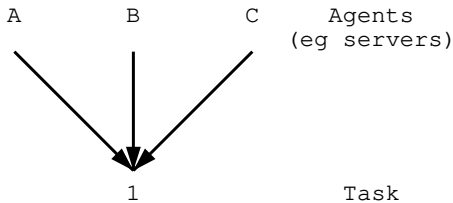}}}
  \caption{This figure shows the task labeled 1 being performed by agents A, B, and C, thus two of these agents are redundant. An example in a network situation would be load-balancing a web server, where any one of three servers can server a particular site to a client. Note that there is an extra cost associated with this redundancy, so in this example we are paying for three servers instead of one. However the system is robust, so if one of the servers is busy or breaks down, then the task (such as serving a web site) can still be performed.}
  \label{redundancy}
\end{figure}

Redundancy is where one task or outcome is determined by more than one agent (assuming the independence of agents, either acting independently or together).  It has the advantage of conferring robustness (integrity) upon the system, because if one agent were to fail, others are able to perform the task.  However, redundancy may be costly, as the overlapping of agents may be inefficient or wasteful.  Despite this, in some systems the wastage may be justified if the task or outcome is so important that the system will fail in its absence, and therefore be selectively disadvantaged. Figure~\ref{redundancy} shows an example of a redundant system.

\begin{figure}[hptb]
  \centering{\resizebox{6cm}{!}{\includegraphics{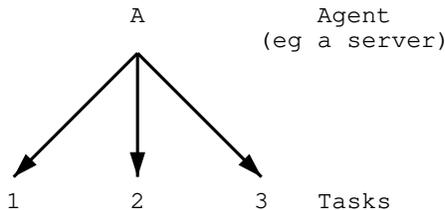}}}
  \caption{This figure shows a single agent, A, performing multiple tasks, labeled 1, 2 and 3. An example in a network situation could be a server handling multiple client requests, such as sending email to client 1 while sending a web page to client 2. While this is cost effective, it lacks the robustness to failure of a redundant system as shown in Figure~\ref{redundancy}.}
  \label{pleiotropy}
\end{figure}

The opposite of redundancy is pleiotropy, where one agent may perform many tasks.  This has a number of distinct advantages.  It is efficient, and allows for spatial and temporal flexibility.  Its major cost is that it is dependent upon the history of the system, that is, any given agent may only be working under particular conditions which have certain constraints imposed upon it by the peculiar evolutionary history of that system.  Despite this, both temporal and spatial pleiotropy may exist, where a given agent can perform qualitatively different tasks, as well as perform the same task (or different tasks) at different times.  As pleiotropy enables efficiency, it therefore confers selective advantages.  How pleoitropic an agent is will depend upon the context in which it is working.  
Some agents may in and of themselves be highly non-specific, and so the outcome is defined by the context in which the agent works, for example, the effect of nitric oxide (NO) is defined not by NO itself, but the molecular context in which NO is bound. At the other extreme, some outcomes may be the result of highly specific interactions, for example protein -- protein binding (receptor - ligand interactions). A simple pleiotropy example is shown in Figure~\ref{pleiotropy}.

\begin{figure}[hptb]
  \centering{\resizebox{8cm}{!}{\includegraphics{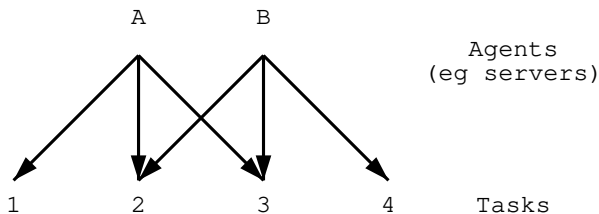}}}
  \caption{This figure shows two agents, both of which perform multiple tasks. An example in a network situation would be two servers, both providing the same email and web services to a number of clients. This system has robustness as if one server fails, both email and web services can still be provided. It also minimizes cost, as it would cost the same as two servers with one providing just email services and the other hosting a web site.}
  \label{both}
\end{figure}

What happens if you combine the two? When both pleiotropy and redundancy are combined, the system possesses properties that it otherwise lacks when pleiotropy or redundancy exist on their own.  The advantages include an increase in the robustness of the system due to the redundancy build into it.  It is more efficient due to the pleiotropy.  The system becomes inherently more flexible and the costs of redundancy are offset by the increase in efficiency due to the presence of pleiotropy. An example of a system with a combination of pleiotropy and redundancy is shown in Figure~\ref{both}.
\section{METHODS}
In this section we describe the structure and representation of the network, the details of the genetic algorithm used, the initialization and parameters used for designing the network using the genetic algorithm, and the fitness function. We developed a graphical user interface (GUI) for running the genetic algorithm, to allow for easy user modification of the network parameters, this also allows one to watch the evolution of the network.
\subsection{Network structure}
The network consists of a set of servers, a set of clients (which can also function as routers), and a set of links between those various nodes. A graph data structure is used to represent the network, with each node (client or server) in the graph having the following properties:
\begin{itemize}
\item node label, ``C'' for a client (including routers) or ``S'' for a server
\item node ID, which also serves as a grid reference of the node for display in a GUI
\item node failure rate, a value between zero and one giving the probability of failure per time step
\item current state, working or non-working
\item number of time steps since failure, zero if working
\item details of the inbound and outbound network connections.
\end{itemize}
The edges, the links in the network, have the following properties:
\begin{itemize}
\item link label indicating whether the link is a link between clients (including routers) or between a client or router to a server
\item edge ID, which also serves as a pair of grid references for display in a GUI
\item edge failure rate, a value between zero and one giving the probability of failure
\item current state, working or non-working
\item number of time steps since failure, zero if working.
\end{itemize}
\subsection{Network construction and maintenance}
\begin{figure}
 \centering{\resizebox{9.1cm}{!}{\includegraphics{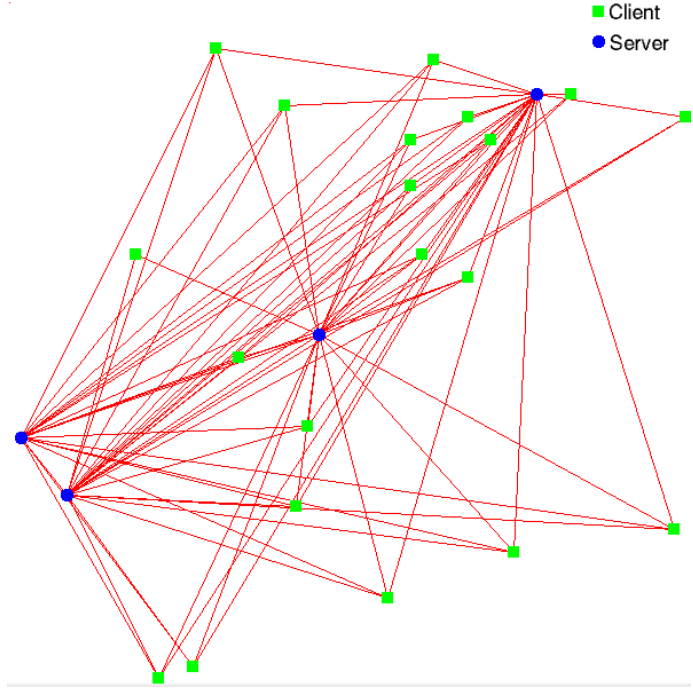}}}
  \caption{This figure shows an example of an evolved network, with clients and servers and a set of links between them. The clients and servers have been positioned at random, with a minimum spacing to avoid clutter.}
\label{network}
\end{figure}

We initially start with a set of clients ($\mathcal{C}$) and servers ($\mathcal{S}$), with no links. The positions of the clients and servers are set at random, with a minimum spacing between them.
Each client $i \in \mathcal{C}$ is assigned a traffic value, $T_{i}$, at random ($0<T_{i}<T_{\mathrm{max}}$), which indicates the amount of traffic requested by the client that is to be transmitted across the network. Each server $j \in \mathcal{S}$ has a fixed amount $T_{s}$ of traffic it can serve. 
We define a utilization parameter,
\begin{equation}
U=\frac{\displaystyle \sum_{i \in \mathcal{C}} T_{i}}{\norm{\mathcal{S}}T_{s}},
\label{utilization}
\end{equation}
describing how well the servers are able to deliver their available load to the clients.
If the utilization is less than $0.75$, then more links are added at random to carry the extra server capacity to clients. If, on the other hand, the utilization is greater than $0.85$ then either links are removed (reducing the amount of traffic that is able to be requested from servers) or more servers are added. The network thus evolves by starting without any connections, and through mutations including:
\begin{itemize}
\item adding links to increase $U$
\item removing links to decrease $U$
\item adding servers to decrease $U$
\item links failing
\item links being repaired.
\end{itemize}
An example of an evolved network is shown in Figure~\ref{network}.

Having established a network, we then need to measure its fitness.
\subsection{Fitness and cost functions}
The aim is to find an optimal network, which minimizes cost ($P$) and maximizes reliability ($R$). With this in mind, we define our fitness function, $F$, to be:
\begin{equation}
F=\frac{R}{P},
\label{fitness}
\end{equation}
where the cost function, $P$, is defined to be the total sum of all the edge lengths of the network. Each node is assigned an $x$ and $y$-coordinate, therefore the edge length, is the length of the straight line connecting the 2 nodes. The reliability function, $R$, is defined as the probability that a connection can be made between any two points. In order to calculate this probability, we randomly pick $N$ pairs of points in the network. For each pair picked we check whether a path exists between those two points. We compute $R$ from the number of pairs for which a path exists divided by $N$, the total number of pairs examined.
\subsection{Redundancy and pleiotropy functions}
We define the overall measure of redundancy for the whole network as
\begin{equation}
D=\frac{\displaystyle \sum_{i\in\mathcal{C}} O_{i}}{\norm{\mathcal{S}}},
\label{redundancyeqn}
\end{equation}
where $D$ is the redundancy, $O_{i}$ the out degree or number of links out of client $i \in \mathcal{C}$, and $\mathcal{S}$ the set of servers.
Similarly, the overall measure of pleiotropy is
\begin{equation}
L=\frac{\displaystyle \sum_{i\in\mathcal{S}}I_{i}}{\norm{\mathcal{C}}},
\label{pleiotropyeqn}
\end{equation}
where $L$ is the pleiotropy, $I_{i}$ the in degree or number of links into server $i \in \mathcal{S}$, and $\mathcal{C}$ the set of clients.
\subsection{Genetic algorithm}
Using the fitness function, we evaluate each network (represented as described above) using the fitness function. We then use two different strategies for evolving the network:
\begin{enumerate}
\item In this strategy we set a variable number of offspring per generation, and at each step pick the single fittest network to reproduce by mutations to create that many offspring, and repeat the process.
\item The second strategy involves creating a fixed number (10) offspring at each step, by picking the fittest two networks and producing five offspring from each. 
\end{enumerate}
We do not consider any mating or crossover between the two due to the complexity of defining a mating operation for networks, and only mutate networks to produce offspring. Crossover would allow a much faster evolution and result in more stability once a fitness plateau is reached~\cite{JHHO}.
\section{RESULTS}
\subsection{Overview}
We used both GA strategies to build a network for a number of different link failure probabilities and repair rates, and evaluated the performance of the strategies, and found the best network parameters to use. Using these network parameters, we then used the best GA strategy to find the best network possible. Note that our GA, since it currently lacks mating and crossover, is more like a Monte-Carlo method for finding the best network than a true GA.
\subsection{Varying failure probability}
\begin{figure}[htbp]
    \centering\mbox{
        	\subfigure[This figure shows the reliability measure for GA strategy one for link failure probabilities of low (0.001) and high (0.01)]{
            \label{linkfail:e1rel}
            \includegraphics[width=5.1cm]{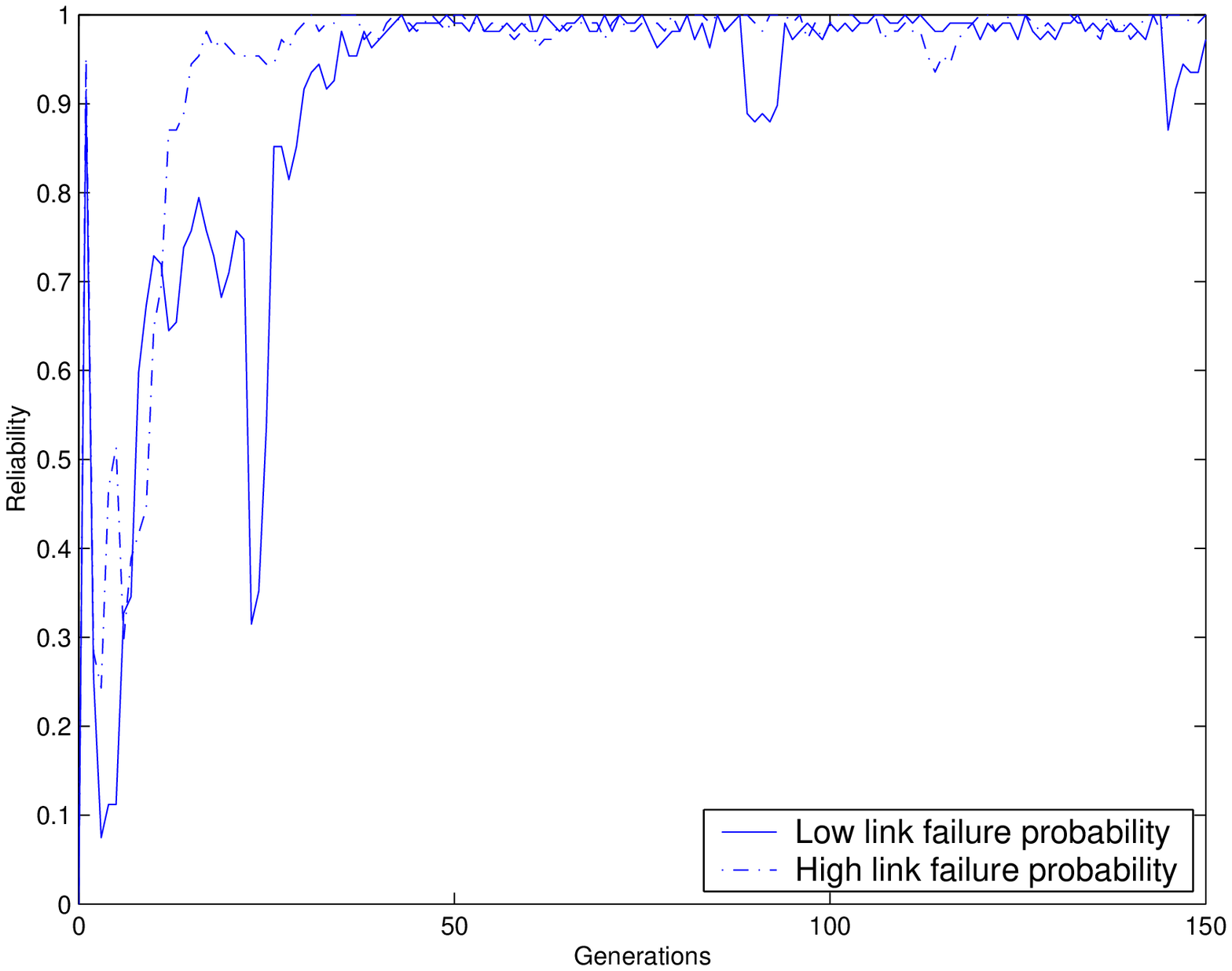}
        }
        	\subfigure[This figure shows the cost measure for GA strategy one for link failure probabilities of low (0.001) and high (0.01)]{
            \label{linkfail:e1cost}
            \includegraphics[width=5.1cm]{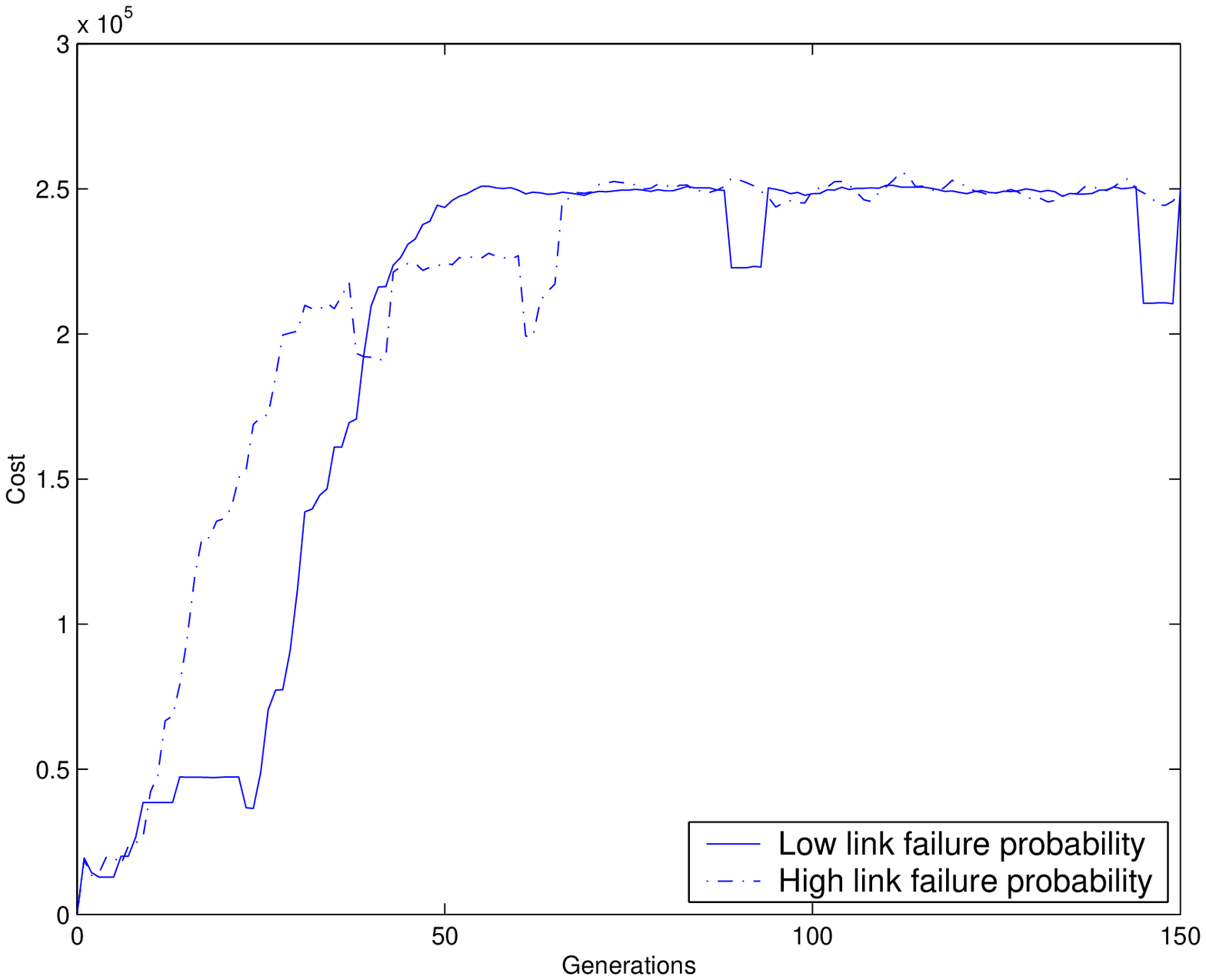}
        }
        \subfigure[This figure shows the redundancy/pleiotropy measure for GA strategy one for link failure probabilities of low (0.001) and high (0.01)]{
            \label{linkfail:e1ratio}
            \includegraphics[width=5.1cm]{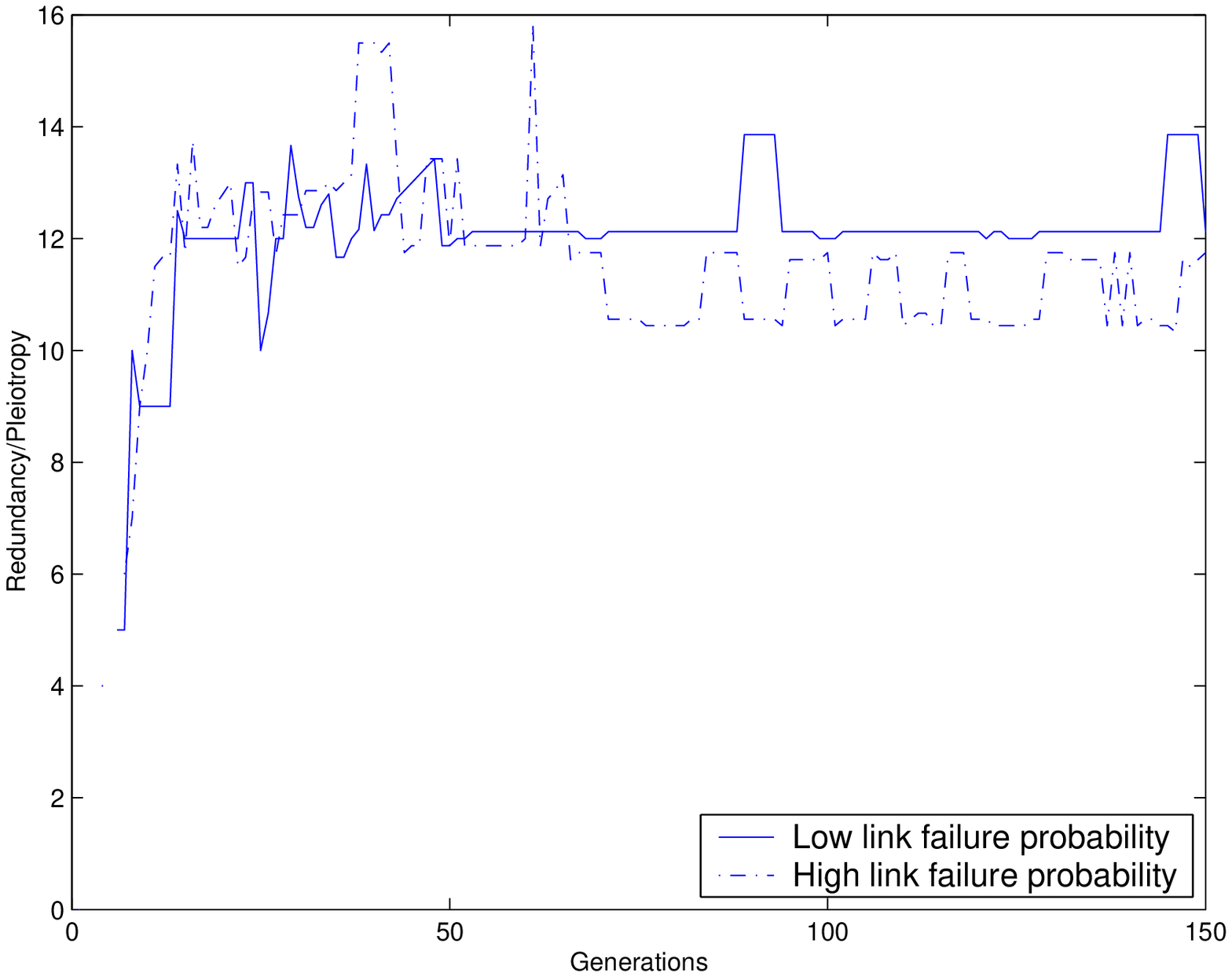}
        }}\quad\mbox{
        \subfigure[This figure shows the reliability measure for GA strategy two for link failure probabilities of low (0.001) and high (0.01)]{
            \label{linkfail:e2rel}
            \includegraphics[width=5.1cm]{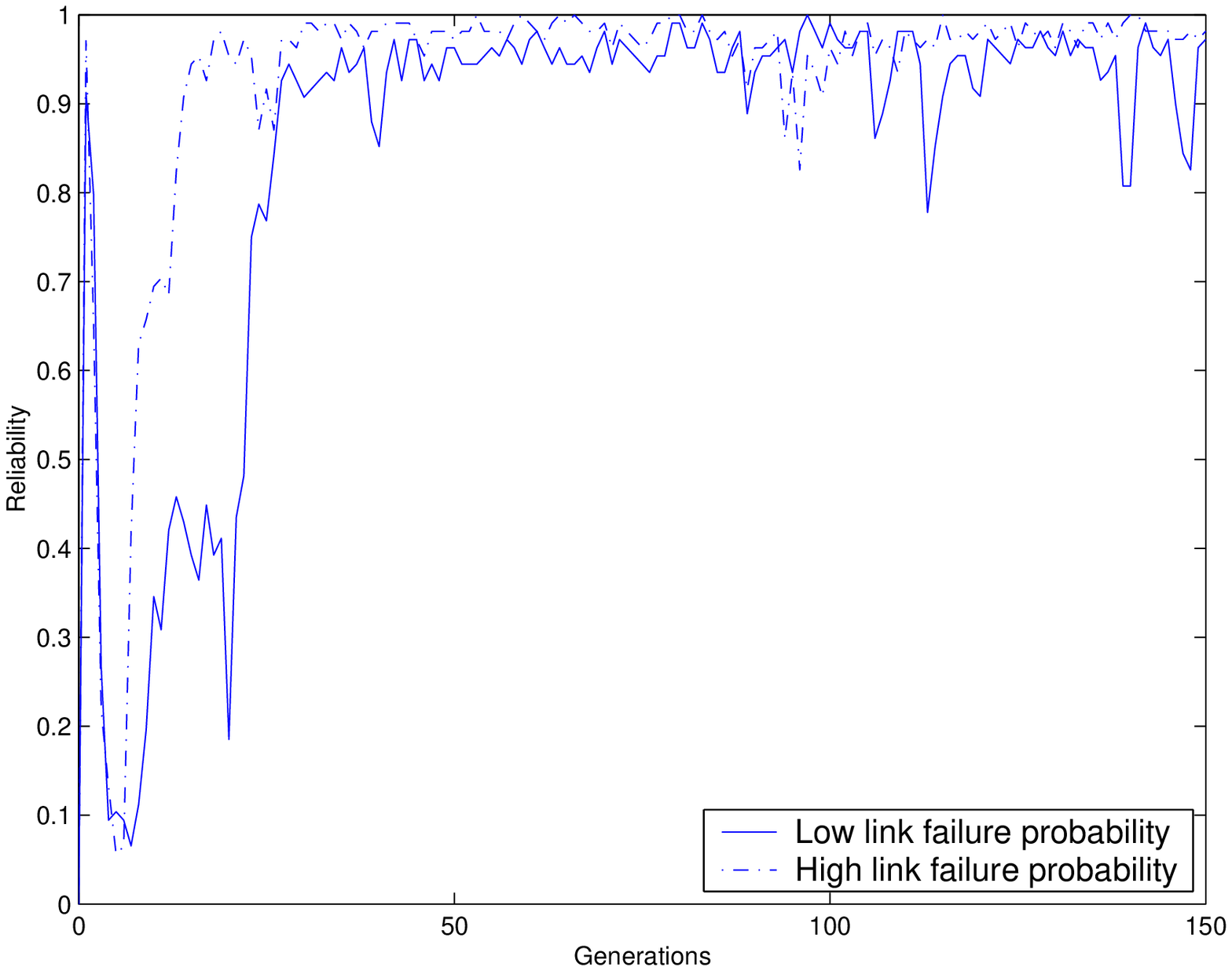}
        }
	\subfigure[This figure shows the cost measure for GA strategy two for link failure probabilities of low (0.001) and high (0.01)]{
            \label{linkfail:e2cost}
            \includegraphics[width=5.1cm]{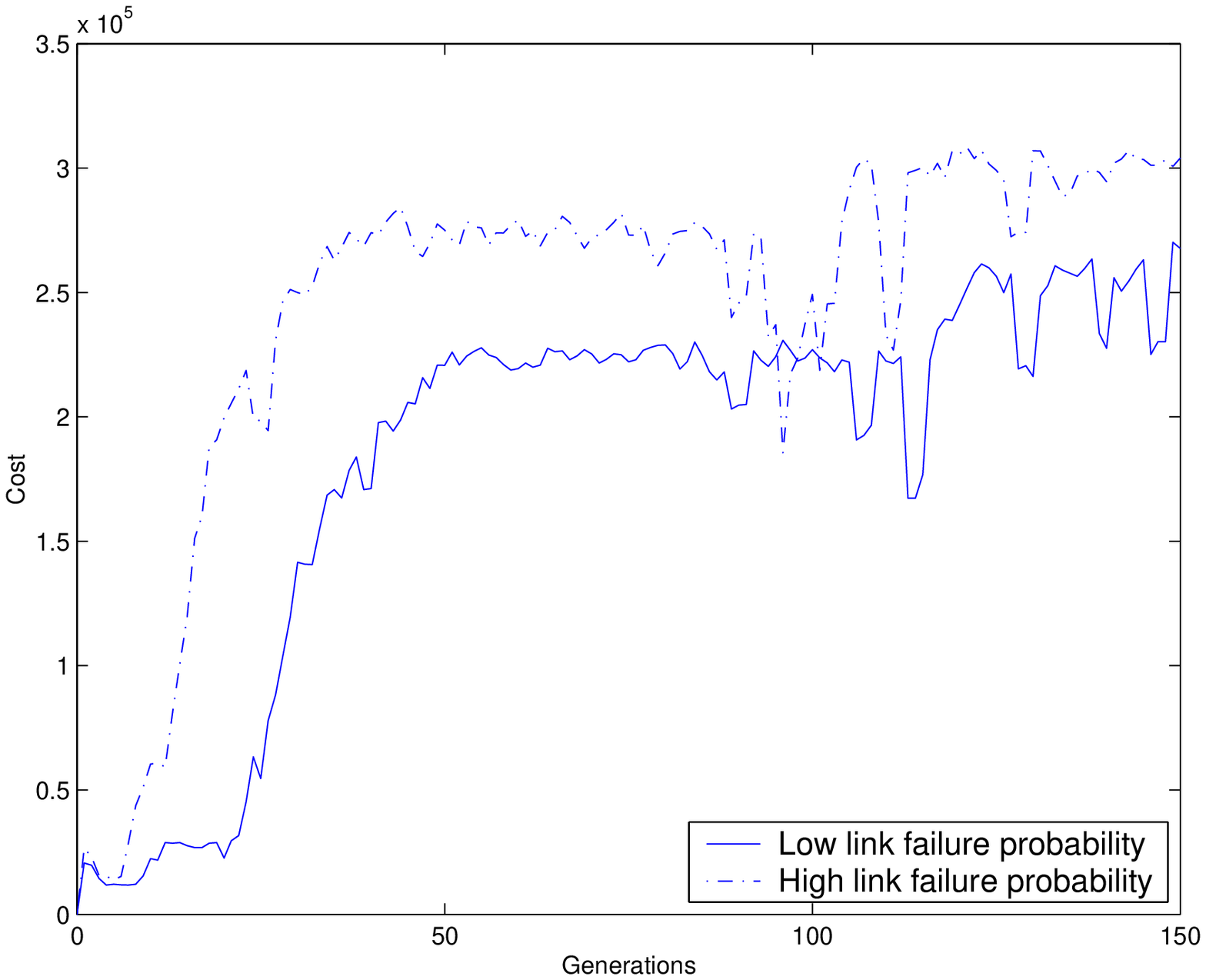}
        }

	\subfigure[This figure shows the redundancy/pleiotropy measure for GA strategy two for link failure probabilities of low (0.001) and high (0.01)]{
            \label{linkfail:e2ratio}
            \includegraphics[width=5.1cm]{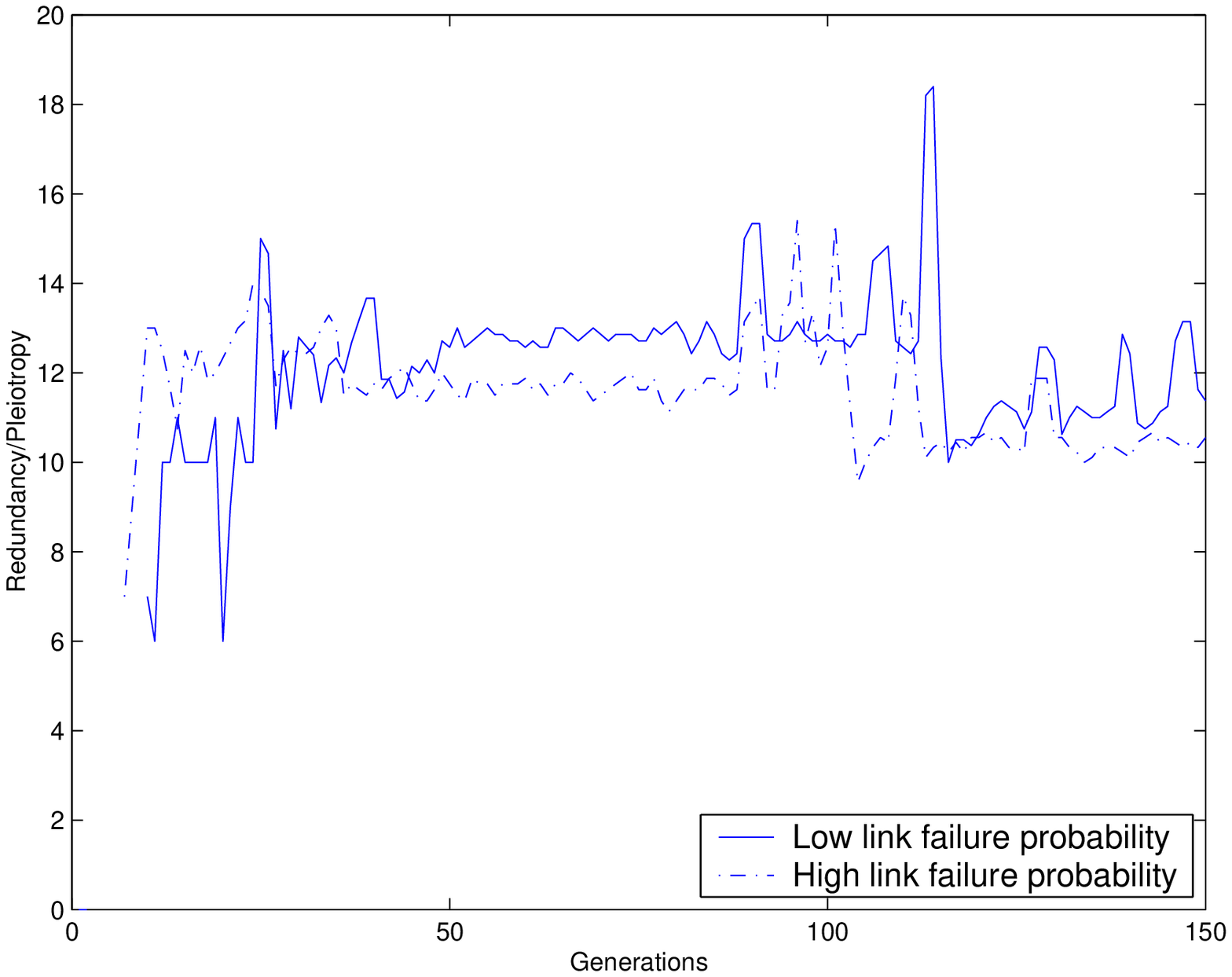}
        }}
    \caption{These graphs show the cost, reliability, and redundancy/pleiotropy functions for both GA strategies and for two link failure probabilities, 0.01 and 0.001. Initial spikes in the graph are caused by sampling error when there are only a few links in the network with which to calculate the measures.}
\label{linkfail}
\end{figure}

We tested the both GA evolution strategies with the link failure probability set to two values, $0.01$ per time step, representing a reasonable failure probability given regular network construction, and $0.001$, representing a low failure probability given a high quality network construction. Figure~\ref{linkfail} shows the evolution of a solution to each of these link failure probabilities using both GA strategies, with an optimal solution being reached in about $50$ generations, after that the random mutations and random link additions or removals result in noise about the optimum solution, as the GA has reached a fitness plateau. The 50 generation mark was common to all our results, thus our tables show the mean and standard deviation of only the network generations occurring after generation $50$. Using the first GA strategy results in a higher network reliability and lower cost, because it allows more variation at each iteration and can thus climb to a higher fitness (reliability/cost) plateau.
\begin{table}[htbp]
\centering\footnotesize
\caption{This table shows the mean and standard deviation, over generations 50-150 of the GA, of the cost and reliability functions for both GA strategies, and for link failure probabilities $0.01$ and $0.001$. We use generations 50-150 as by generation 50 the GA has optimized the network to a fitness plateau, as evident in Figure~\ref{linkfail}. Note that our measure of reliability has a maximum achievable value of one. The results indicate no significant difference between the reliabilities of the two failure probabilities, however there is some significant difference between the costs, with strategy two performing better for a lower failure probability (t-test, confidence level of 95\%)}
\begin{tabular}{|c|c|c|}\hline
& {\bf failure prob. = 0.01} & {\bf failure prob. = 0.001}\\\hline
{\bf Mean reliability, strategy one} & $0.988$ & $0.979$\\\hline
{\bf SD reliability, strategy one} & $0.012$ & $0.028$\\\hline
{\bf Mean reliability, strategy two} & $0.974$ & $0.948$ \\\hline
{\bf SD reliability, strategy two} &  $0.025$ & $0.040$\\\hline
{\bf Mean cost (\$ '000s), strategy one} & $244.7 $ & $246.1 $\\\hline
{\bf SD cost (\$ '000s), strategy one} & $11.7$ & $10.0$\\\hline
{\bf Mean cost (\$ '000s), strategy two} & $277.4 $ & $228.3 $\\\hline
{\bf SD cost (\$ '000s), strategy two} & $24.0$ & $19.2$\\\hline
\end{tabular}
\label{linkfailtable}
\end{table}
\subsection{Varying repair rate}
\begin{figure}[hptb]
    \centering\mbox{
        \subfigure[This figure shows the reliability measure for GA strategy one for varying repair rates.]{
            \label{repair:e1rel}
            \includegraphics[width=5.1cm]{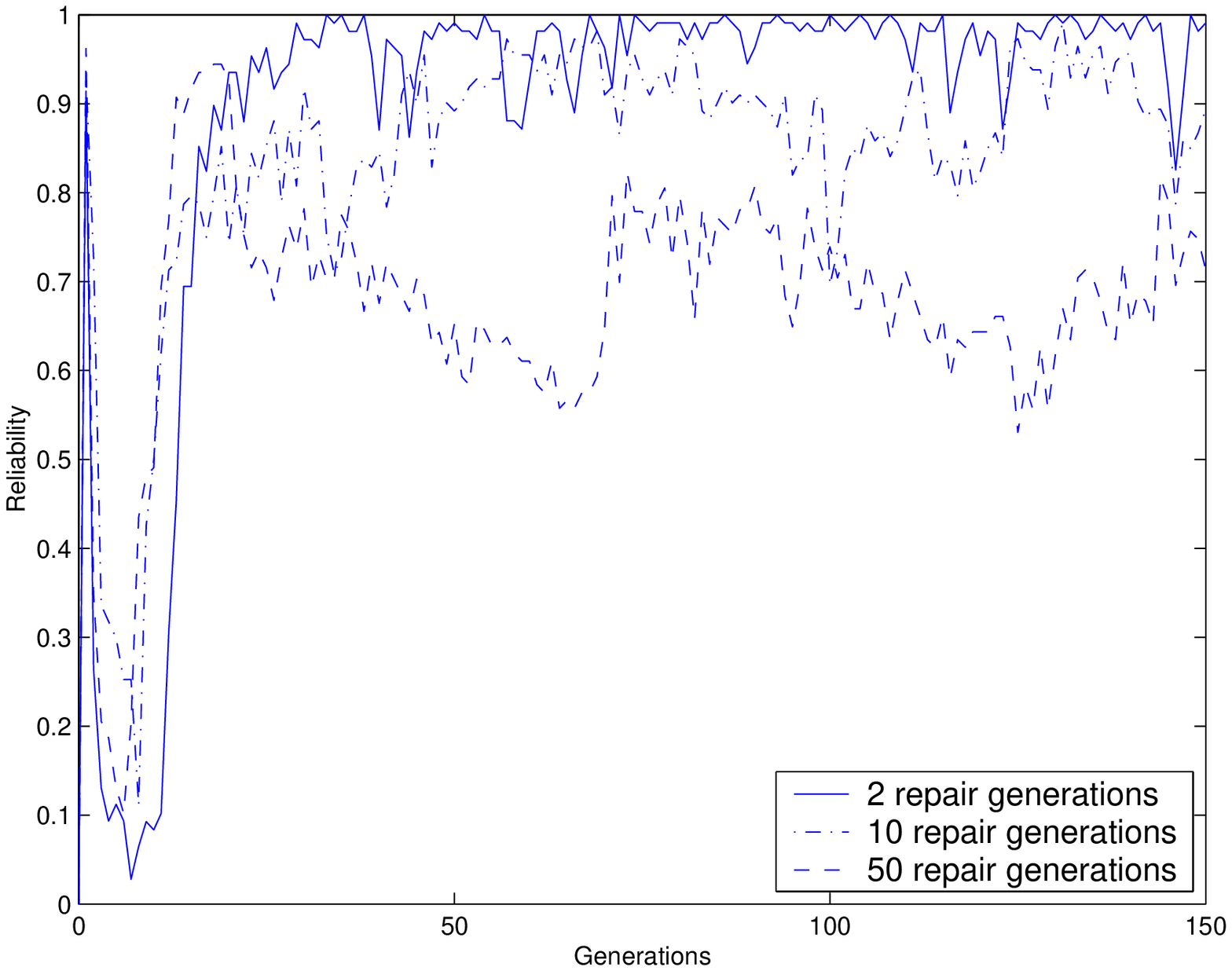}
        }
        \subfigure[This figure shows the cost measure for GA strategy one for varying repair rates.]{
            \label{repair:e1cost}
            \includegraphics[width=5.1cm]{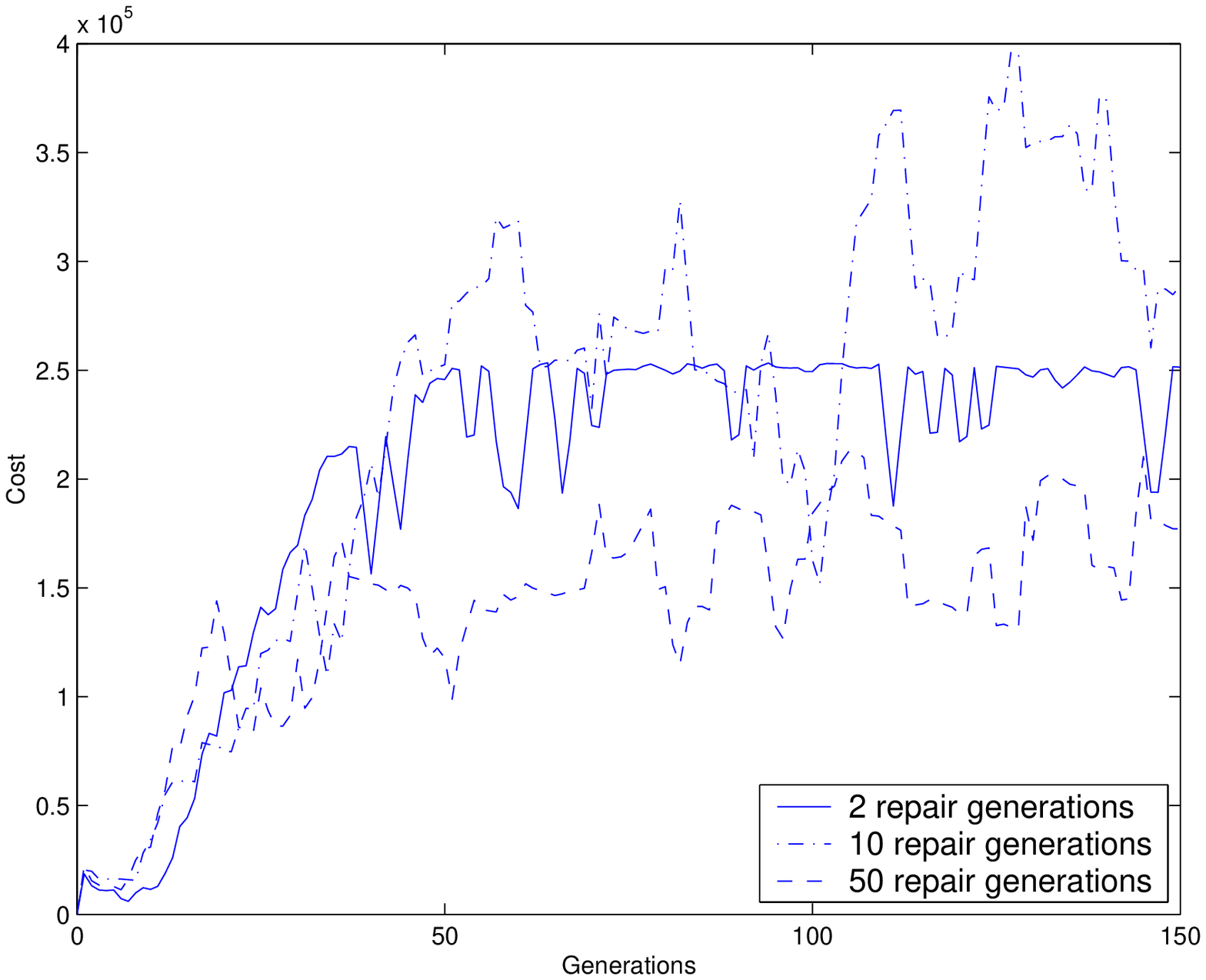}
        }
        	\subfigure[This figure shows the redundancy/pleiotropy measure for GA strategy one for varying repair rates.]{
            \label{repair:e1ratiot}
            \includegraphics[width=5.1cm]{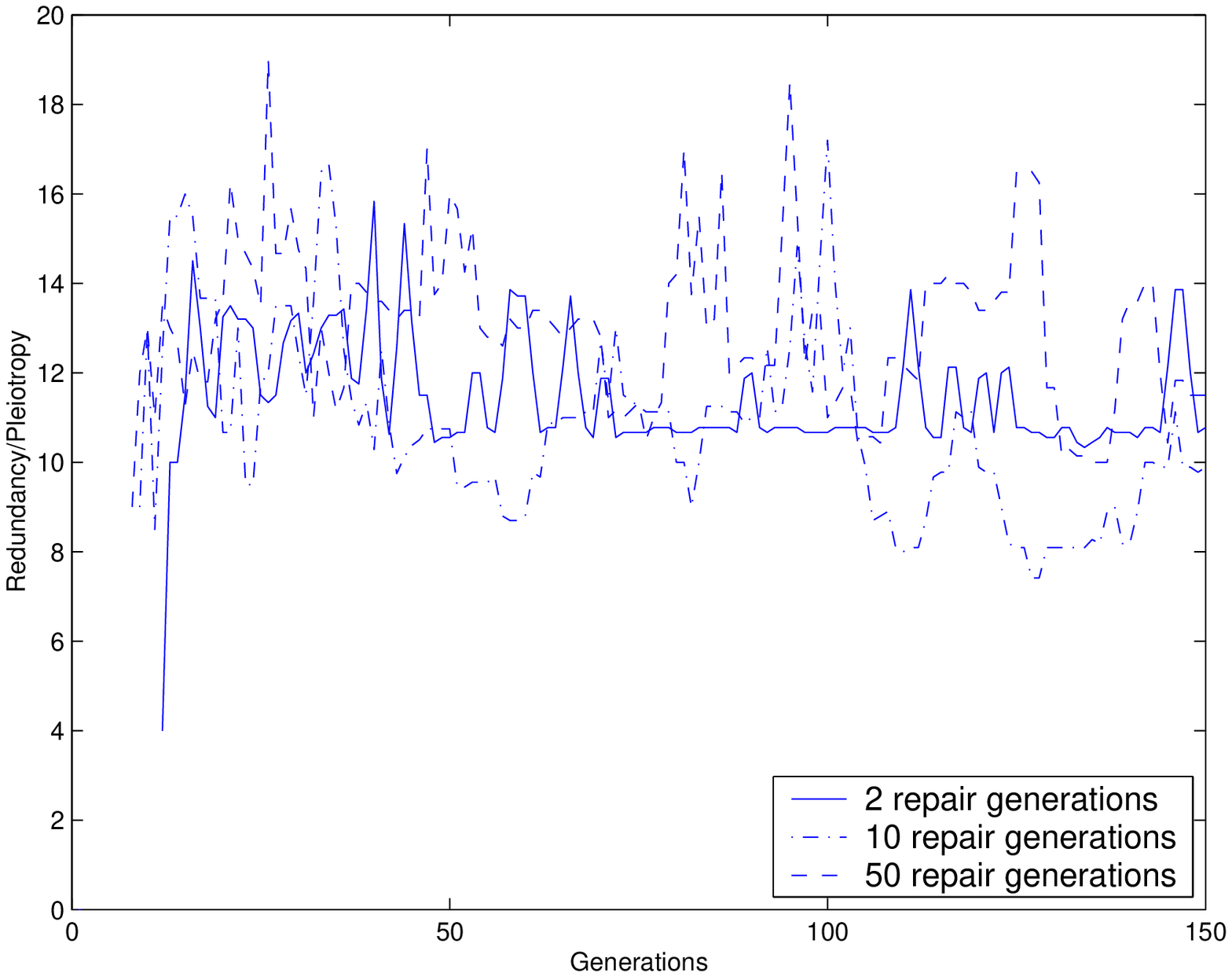}
        }}\quad\mbox{
        \subfigure[This figure shows the reliability measure for GA strategy two for varying repair rates.]{
            \label{repair:e2rel}
            \includegraphics[width=5.1cm]{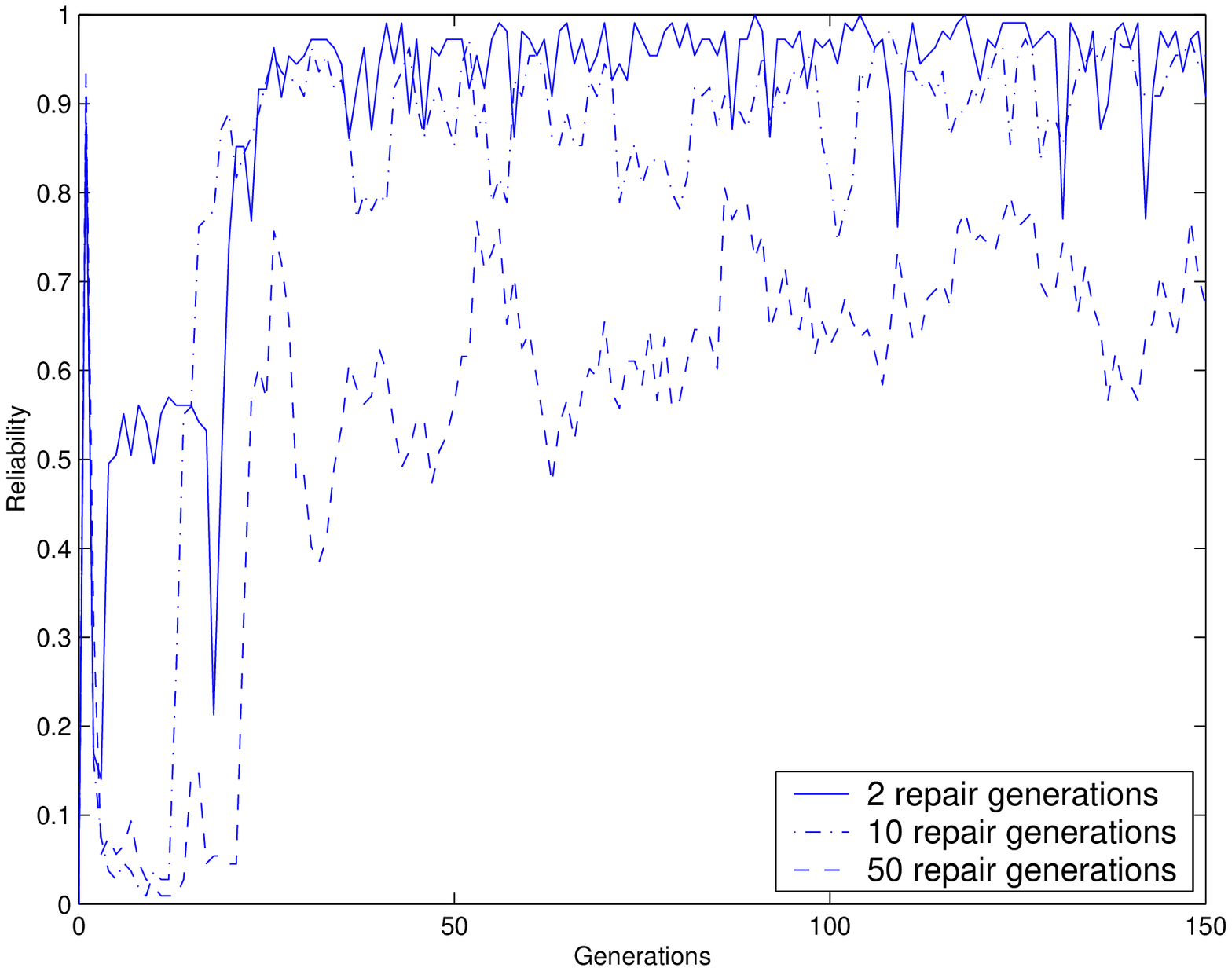}
        }
	\subfigure[This figure shows the cost measure for GA strategy two for varying repair rates.]{
            \label{repair:e2cost}
            \includegraphics[width=5.1cm]{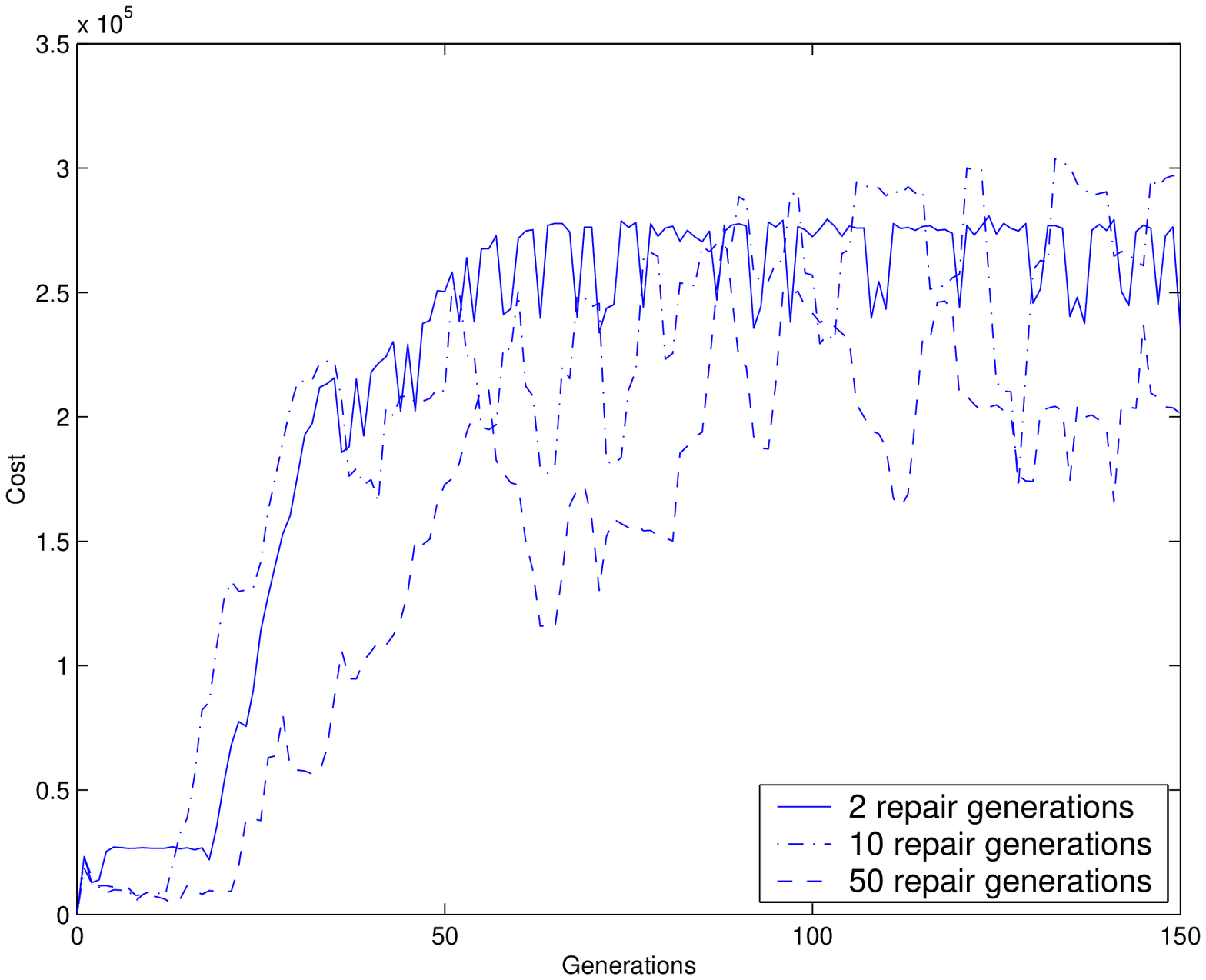}
        }

	\subfigure[This figure shows the redundancy/pleiotropy measure for GA strategy two for varying repair rates.]{
            \label{repair:e2ratio}
            \includegraphics[width=5.1cm]{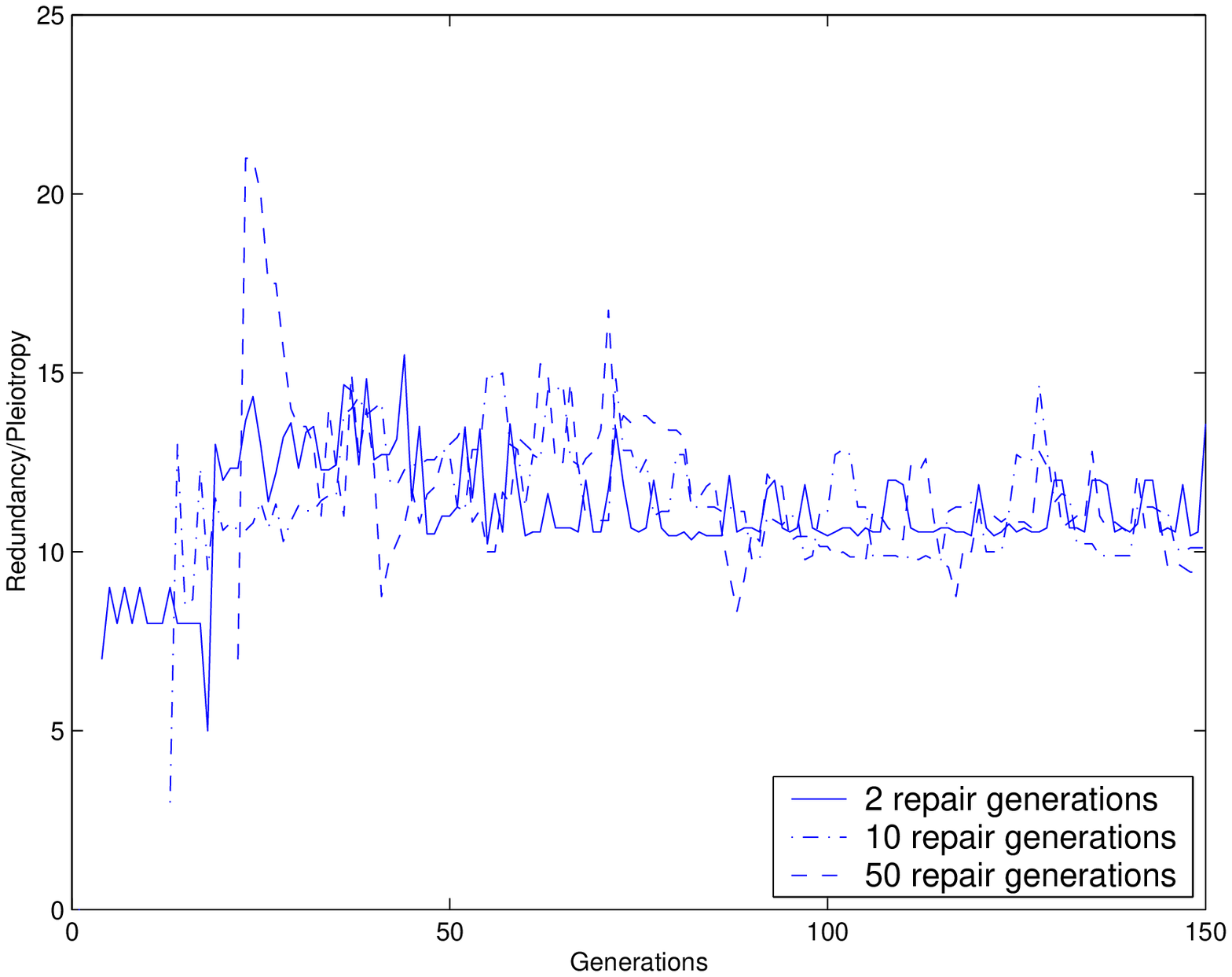}
        }}
    \caption{These graphs show the cost, reliability, and redundancy/pleiotropy functions for both GA strategies and for varying repair rates. The repair process fixes links after 2, 10 or 50 generations as selected by the user of the software. Initial spikes in the graph are caused by sampling error when there are only a few links in the network with which to calculate the measures.}
\label{repair}
\end{figure}

We tested the both GA evolution strategies with the link repair time set to repair times of 2, 10, and 50 generations, representing ideal, average, and worst case repair processes. Figure~\ref{repair} shows the evolution of a solution to each of these repair times using both GA strategies, again we find an optimal solution being reached in $50$ generations, after that the random mutations and random link additions or removals result in noise about the optimum solution. Using the first GA strategy results in a higher network reliability and lower cost, because it allows more variation at each iteration and can thus climb to a higher fitness (reliability/cost) plateau.
\begin{table}[hptb]
\centering\footnotesize
\caption{This table shows the mean and standard deviation, over generations 50-150 of the GA, of the cost and reliability functions for both GA strategies, and varying repair rates (2, 10, and 50 generations).
We use generations 50-150 as by generation 50 the GA has optimized the network to a fitness plateau, as evident in Figure~\ref{repair}. Note that our measure of reliability has a maximum achievable value of one. Here there is a significant difference as we go from 2 to 50 generations (t-test, confidence level of 95\%).}
\begin{tabular}{|c|c|c|c|}\hline
& {\bf 2 generations} & {\bf 10 generations} & {\bf 50 generations}\\\hline
{\bf Mean reliability, strategy one} &$0.973$ & $0.902$ & $0.679$\\\hline
{\bf SD reliability, strategy one} & $0.034$ & $0.054$ & $0.072$\\\hline
{\bf Mean reliability, strategy two} & $0.956$ & $0.901$ & $0.661$\\\hline
{\bf SD reliability, strategy two} & $0.045$ & $0.055$ & $0.073$\\\hline
{\bf Mean cost (\$ '000s), strategy one} & $240.5$ & $288.3$ & $162.1$\\\hline
{\bf SD cost (\$ '000s), strategy one} & $17.8$ & $51.8$ & $25.5$\\\hline
{\bf Mean cost (\$ '000s), strategy two} & $266.4$ & $253.1$ & $194.0$\\\hline
{\bf SD cost (\$ '000s), strategy two} & $14.8$ & $35.2$ & $33.5$\\\hline
\end{tabular}
\label{repairtable}
\end{table}
\subsection{Number of offspring for GA strategy one}
Here we considered what happens if we change the number of offspring produced at each step of GA strategy one. The results are shown in Table~\ref{offspring}, indicating that 10 offspring produces the fittest network. 
\begin{table}[hptb]
\centering\footnotesize
\caption{This table shows the mean and standard deviation, over generations 50-150 of the GA, of the cost and reliability functions for GA strategy one, and varying number of offspring (10, 20, and 50).
We use generations 50-150 as by generation 50 the GA has optimized the network to a fitness plateau (not shown). Note that our measure of reliability has a maximum achievable value of one.}
\begin{tabular}{|c|c|c|c|}\hline
& {\bf 10 offspring} & {\bf 20 offspring} & {\bf 50 offspring}\\\hline
{\bf Mean reliability} & $0.957$ & $0.910$ & $0.901$\\\hline
{\bf SD reliability} & $0.039$ & $0.050$ & $0.060$\\\hline
{\bf Mean cost (\$ '000s} & $254.1$ & $274.7$ & $249.4$\\\hline
{\bf SD cost (\$ '000s)} & $35.7$ & $34.6$ & $36.2$\\\hline
\end{tabular}
\label{offspring}
\end{table}
\section{CONCLUSIONS}
Genetic algorithms rapidly converge on optimal real-world network design solutions, where both cost and reliability are important. Pleiotropy helps reduce the cost, and redundancy improves the reliability and network traffic flows. For the parameters and methods considered, we found that strategy one found the best networks (in terms of $R/C$, so lowest cost and highest reliability) for a range of link failure and repair rates. This was due to more variance at each generation, allowing the network to climb to a higher fitness plateau. Furthermore, we found that strategy one works best when 10 offspring are produced from the fittest network at each generation, higher numbers of offspring tend to add too much variance to the process.

In future work we propose using the GA on not just the network layout but to include the failure and repair rates in the ``genome''. A basic implementation of crossover would allow for much better network designs~\cite{JHHO}. Making the fitness function as a summation of a set of local fitness functions for individual clients would provide a faster and more accurate way of measuring the fitness.
\acknowledgments
We gratefully acknowledge funding from The University of Adelaide. Erin OÕNeill would like to thank Chris Bryant for helpful discussions of this work. Erin O'Neill is supported by ARC Grant No.~A00103779.
\bibliography{phd}   
\bibliographystyle{spiebib}   
\end{document}